\documentclass[10pt,twocolumn,letterpaper]{article}

\usepackage{cvpr}
\usepackage{times}
\usepackage{epsfig}
\usepackage{graphicx}
\usepackage{amsmath}
\usepackage{amssymb}
\usepackage{algorithm}
\usepackage{algpseudocode}

\usepackage[breaklinks=true,bookmarks=false]{hyperref}

\cvprfinalcopy 

\def\cvprPaperID{1317} 

\begin{document}
\graphicspath{{figures/}}
\title{Low-Latency Video Semantic Segmentation}

\author{Yule Li$^1$\thanks{This work is done when Yule Li is intern at
CUHK Multimedia Lab }, \qquad \qquad Jianping Shi$^2$,\qquad \qquad Dahua Lin$^3$\\
$^1$Key Laboratory of Intelligent Information Processing of Chinese Academy of Sciences (CAS),\\
Institute of Computing Technology, Chinese Academy of Sciences, Beijing, 100190, China\\
$^2$SenseTime Research, $^3$Department of Information Engineering, The Chinese University of Hong Kong \\
{\tt\small yule.li@vipl.ict.ac.cn, shijianping@sensetime.com, dhlin@ie.cuhk.edu.hk}
}

\maketitle


\begin{abstract}
Recent years have seen remarkable progress in semantic segmentation.
Yet, it remains a challenging task to apply segmentation
techniques to video-based applications.
Specifically, the high throughput of video streams,
the sheer cost of running fully convolutional networks, together
with the low-latency requirements in many real-world applications,
\eg~autonomous driving, present a significant challenge to the
design of the video segmentation framework.
To tackle this combined challenge,
we develop a framework for video semantic segmentation,
which incorporates two novel components:
(1) a feature propagation module that adaptively fuses features over time
via spatially variant convolution, thus reducing the cost of per-frame
computation; and
(2) an adaptive scheduler that dynamically allocate computation
based on accuracy prediction.
Both components work together to ensure low latency while maintaining high
segmentation quality.
On both Cityscapes and CamVid,
the proposed framework obtained competitive performance compared to
the state of the art,
while substantially reducing the latency,
from $360$ ms to $119$ ms.
\end{abstract}



\section{Introduction}
\label{sec:intro}




Semantic segmentation, a task to divide observed scenes into semantic regions,
has been an active research topic in computer vision.
In recent years, the advances in
deep learning~\cite{krizhevsky2012imagenet,simonyan2014very,he2015deep} and
in particular the development of Fully Convolutional Network (FCN)~\cite{long2015fully}
have brought the performance of this task to a new level.
Yet, many existing methods for semantic segmentation were devised for parsing
images~\cite{long2015fully,chen2014semantic,liu2015semantic,noh2015learning,zhao2016pyramid}.
How to extend the success of segmentation techniques to video-based
applications (\eg~robotics, autonomous driving, and surveillance) remains
a challenging question.


The challenges of video-based semantic segmentation consist in two aspects.
On one hand, videos usually involve significantly larger volume of data
compared to images. Particularly, a video typically contains $15$ to $30$
frames per second. Hence, analyzing videos requires much more computing
resources.
On the other hand, many real-world systems that need video
segmentation, \eg~autonomous driving, have strict requirments
on the \emph{latency} of response, thus making the problem even more challenging.

\begin{figure}[t]
    \centering
    \includegraphics[width = 0.92\linewidth]{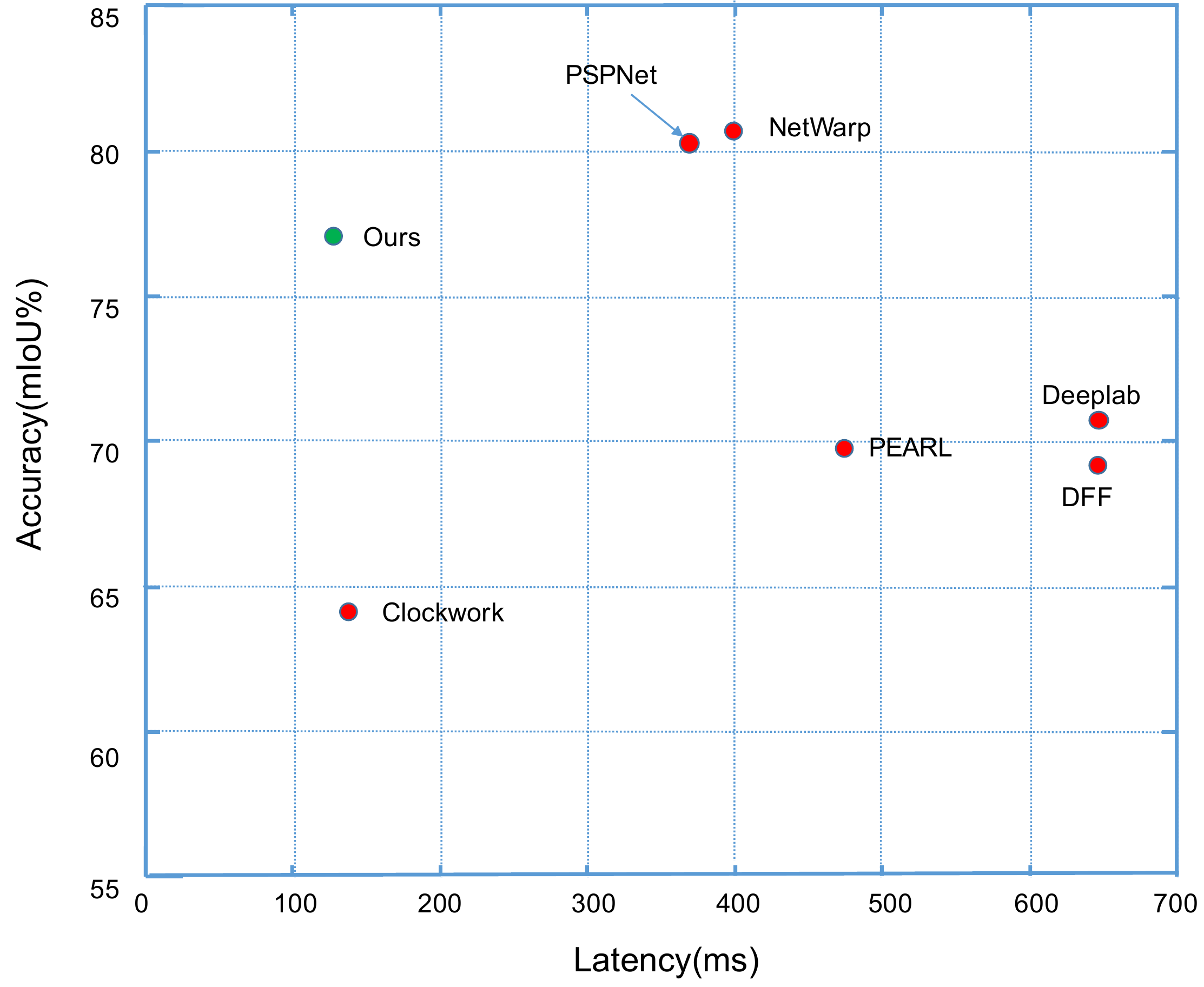}
    \caption{\small
        Latency and mIoU performance on Cityscapes~\cite{cordts2016cityscapes} dataset.
        Methods involved are NetWarp~\cite{gadde2017semantic},
        PSPNet~\cite{zhao2016pyramid}, Deeplab~\cite{chen2016deeplab},
        PEARL~\cite{jin2016video}, DFF~\cite{zhu2017deep},
        Clockework~\cite{shelhamer2016clockwork} and Ours.
        Our method achieves the lowest latency while maintaining competitive
        performance.
    }
    \label{fig:latency_acc}
\end{figure}


Previous efforts on video semantic segmentation mainly fall into two classes,
namely \emph{high-level modeling} and \emph{feature-level propagation}.
The former~\cite{fayyaz2016stfcn,nilsson2016semantic} integrates frame-wise analysis via a sequential model. The methods
along this line usually add additional levels on top, and therefore are unable
to reduce the computing cost.
The latter, such as Clockwork Net~\cite{shelhamer2016clockwork} and
Deep Feature Flow~\cite{zhu2017deep}, instead attempts to reuse the features
in preceding frames to accelerate computation.
Such methods were designed to reduce the overall cost in an
\emph{amortized} sense, while neglecting the issue of \emph{latency}.
Moreover, from a technical standpoint, existing methods exploit temporal
correlations by treating all locations \emph{independently} and \emph{uniformly}.
It ignores the different characteristics between smooth regions and boundaries,
and lacks the flexibility of handling complex variations.
Fig.~\ref{fig:latency_acc} compares the performance/latency tradeoffs
of various methods. We can see that previous methods fall short
in either aspect.


Our primary goal in this work is to reduce not only the overall cost but also
the maximum latency, while maintaining competitive performance
in complicated and ever-changing scenarios.
Towards this goal, we explore a new framework.
Here, we adopt the idea of feature sharing, but move beyond the limitations
of previous methods in two important aspects.
(1) We introduce an \emph{adaptive feature propagation} component, which
combines features from preceding frames via
\emph{spatially variant convolution}. By adapting the combination weights
locally, it results in more effective use of previous features and thus
higher segmentation accuracy.
(2) We \emph{adaptively allocate} the keyframes on demand based on
accuracy prediction and incorporate a parallel scheme to coordinate
keyframe computation and feature propagation.
This way not only leads to more efficient use of computational resources
but also reduce the maximum latency.
These components are integrated into a network.


Overall, the contributions of this work lie in three key aspects.
\textbf{(1)} We study the issue of latency, which is often overlooked
in previous work, and present a framework that achieves low-latency
video segmentation.
\textbf{(2)} We introduce two new components: a network using spatially
variant convolution to propagate features adaptively and an adaptive
scheduler to reduce the overall computing cost and ensure low latency.
\textbf{(3)} Experimental results on both Cityscapes and CamVid demonstrate
that our method achieve competitive performance as compared to
the state of the art with remarkably lower latency.


\section{Related Work}
\label{sec:related}



\paragraph{Image Semantic Segmentation}

Semantic segmentation predicts per-pixel semantic labels given the input image.
The Fully Convolutional Network (FCN)~\cite{long2015fully} is a seminal work
on this topic, which replaces fully-connected layers in a
classification network by convolutions to achieve pixel-wise prediction.
Extensions to this formulation mainly follow two directions.
One is to apply
Conditional Random Fields (CRF) or their variants on top of
the CNN models to increase localization
accuracy~\cite{chen2014semantic,liu2015semantic,zheng2015conditional}.
The other direction explores multi-scale architectures to combine
both low-level and high-level
features~\cite{chen2015attention,hariharan2015hypercolumns}.
Various improved designs were proposed in recent years.
Noh \etal~\cite{noh2015learning} proposed to learn a deconvolution network
for segmentation.
Badrinarayanan \etal~\cite{badrinarayanan2015segnet} proposed SegNet,
which adopts an encoder-decoder architecture and leverages max pooling indices
for upsampling.
Paszke \etal~\cite{paszke2016enet} focused on the efficiency and developed ENet,
a highly efficient network for segmentation.
Zhao \etal~\cite{zhao2016pyramid} presented the PSPNet, which
uses pyramid spatial pooling to combine global and local cues.
All these works are purely image-based. Even if applied to videos,
they work on a per-frame basis without considering the temporal relations.


\vspace{-9pt}
\paragraph{Video Semantic Segmentation}

Existing video semantic segmentation methods roughly fall in two categories.
One category is to improve the accuracy by exploiting temporal continuity.
Fayyaz~\etal~\cite{fayyaz2016stfcn} proposed a spatial-temporal LSTM
on per-frame CNN features.
Nilsson and Sminchisescu~\cite{nilsson2016semantic} proposed gated recurrent
units to propagate semantic labels.
Jin~\etal~\cite{jin2016video} proposed to learn discriminative features
by predicting future frames and combine both the predicted results and
current features to parse a frame.
Gadde~\etal~\cite{gadde2017semantic} proposed to combine the features
wrapped from previous frames with flows
and those from the current frame to predict the segmentation.
While these methods improve the segmentation accuracy by exploting
across-frame relations, they are built on per-frame feature computation
and therefore are not able to reduce the computation.

Another category focuses instead on reducing the computing cost.
Clockwork Net~\cite{shelhamer2016clockwork} adapts mutli-stages FCN
and directly reuses the second or third stage features of preceding
frames to save computation.
Whereas the high level features are relatively stable,
such simple replication is not the best practice in general, especially
when signficant changes occur in the scene.
The DFF~\cite{zhu2017deep} propagates the high level feature from the key frame to
current frame by optical flow learned in a flow network~\cite{fischer2015flownet}
and obtains better performance.
Nevertheless, the separate flow network increases the computational
cost; while the per-pixel location transformation by optical flow
may miss the spatial information in the feature field.
For both methods, the key frame selection is crucial to the overall performance.
However, they simply use fixed-interval schedules~\cite{shelhamer2016clockwork,zhu2017deep}
or heuristic thresholding schemes~\cite{shelhamer2016clockwork},
without providing a detailed investigation.
Moreover, while being able to reduce the overall cost, they do not decrease
the maximum latency.



There are also other video segmentation methods but with different settings.
Perazzi~\etal~\cite{perazzi2016benchmark} built a large scale video \emph{object
segmentation} dataset, which concerns about segmenting the foreground objects
and thus are different from our task, parsing the entire scene.
Khoreva~\etal~\cite{khoreva2016learning}
proposed to learn smooth video prediction from static images by combining
the results from previous frames.
Caelles~\etal~\cite{caelles2016one} tackled this problem
in a semi-supervised manner.
Mahasseni~\etal~\cite{mahasseni2017budget} developed
an \emph{offline} method, which relies on a Markov decision process to
select key frames, and propagate the results on key frames to others
via interpolation.
This method needs to traverse the video frames back and forth,
and therefore is not suitable for the online settings discussed in this paper.


\section{Video Segmentation Framework}
\label{sec:frmwork}

%
%

We develop an efficient framework for
video semantic segmentation.
Our goal is to reduce not only the overall computing cost
but also the maximum latency, while maintaining competitive performance.
Specifically, we adopt the basic paradigm of
the state-of-the-art frameworks~\cite{shelhamer2016clockwork,zhu2017deep},
namely, propagating features from key frames to others by exploiting
the strong correlations between adjacent frames.
But we take a significant step further, overcoming the limitations
of previous work with new solutions to two key problems:
(1) \emph{how to select the key frames} and
(2) \emph{how to propagate features across frames}.


\begin{figure*}
    \centering
    \includegraphics[width=0.91\textwidth]{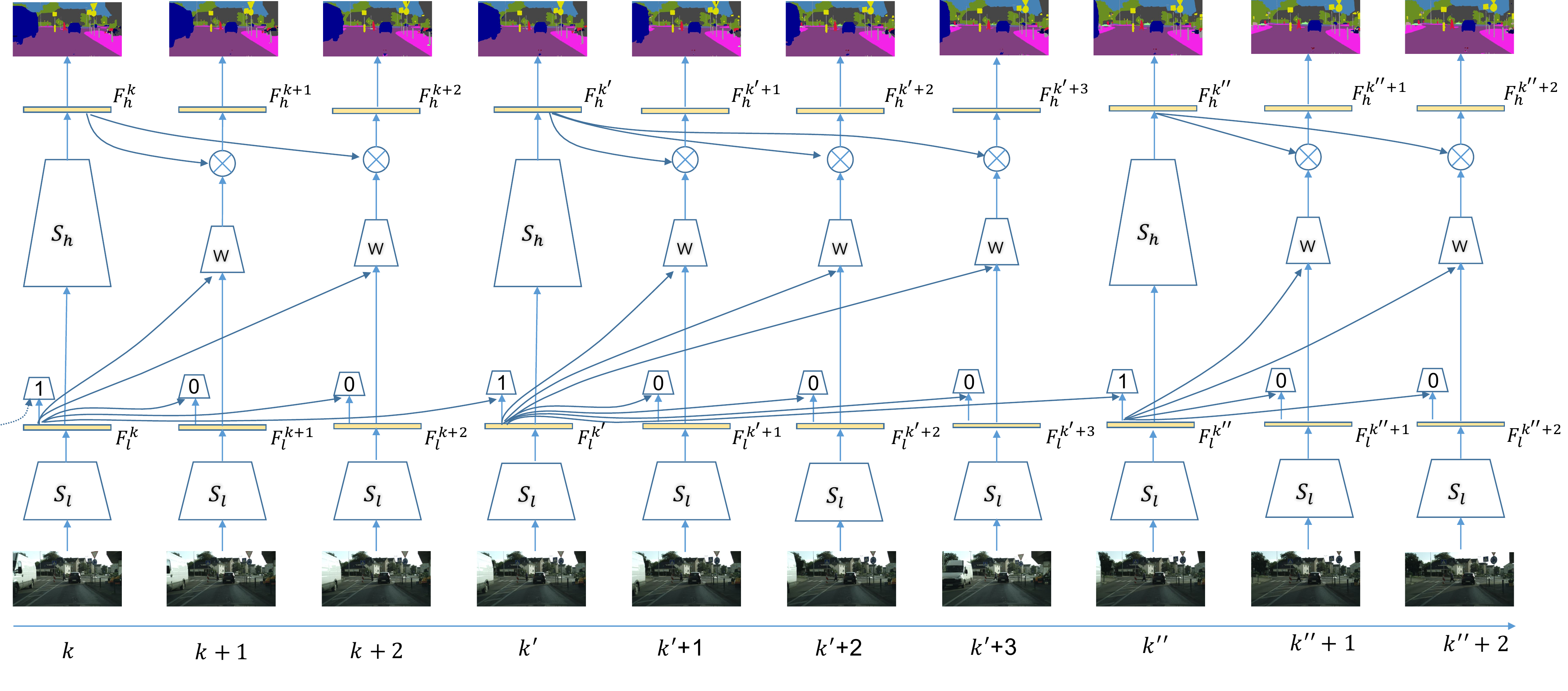}
    \caption{\small
        The overall pipeline.
        At each time step $t$, the lower-part of the CNN $S_l$ first computes
        the low-level features $F_l^t$. Based on both $F_l^k$ (the low-level
        features of the previous key frame) and $F_l^t$, the framework will
        decide whether to set $I^t$ as a new key frame. If yes, the high-level
        features $F_h^t$ will be computed based on the expensive higher-part
        $S_h$; otherwise, they will be derived by propagating from $F_h^k$
        using spatially variant convolution. The high-level features,
        obtained in either way, will be used in predicting semantic labels.
    }
    \label{fig:frmwork}
\end{figure*}

\subsection{Framework Overview}
\label{sub:overview}


Previous works usually select key frames based on
fixed intervals~\cite{zhu2017deep} or simple heuristics~\cite{shelhamer2016clockwork},
and propagate features based on optical flows that are costly
to compute~\cite{zhu2017deep} or CNNs with fixed kernels.
Such methods often lack the capability of handling complex variations
in videos, \eg~the changes in camera motion or scene structures.
In this work, we explore a new idea to tackle these problems --
\emph{taking advantage of low-level features}, \ie~those from lower layers of a CNN.
Specifically, low-level features are inexpensive to obtain,
yet they provide rich information about the characteristics of the underlying frames.
Hence, we may select key frames and propagate features more effectively
by exploiting the information contained in the low-level features,
while maintaining a relatively low computing cost.


Figure~\ref{fig:frmwork} shows the overall pipeline of our framework.
Specifically, we use a deep convolutional network
(ResNet-101~\cite{he2015deep} in our implementation) to extract visual features
from frames. We divide the network into two parts, the lower part $S_l$
and the higher-part $S_h$. The low-level features derived from $S_l$
will be used for selecting key frames and controlling how high-level
features are propagated.

At runtime, to initialize the entire procedure, the framework will feed
the first frame $I^0$ through the entire CNN and obtain both
low-level and high-level features.
At a later time step $t$, it performs the computation \emph{adaptively}.
In particular, it first feeds the corresponding frame $I^t$ to $S_l$,
and computes the low-level features $F_l^t$.
Based on $F_l^t$, it decides whether to treat $I^t$ as a new key frame,
depending on how much it deviates from the previous one.
If the decision is \emph{``yes''}, it will continue to feed $F_l^t$ to
$S_h$ to compute the high-level features $F_h^t$,
and then the segmentation map (via pixel-wise classification).
Otherwise, it will feed $F_l^t$ to a \emph{kernel predictor}, obtain a set
of convolution kernels therefrom, and use them to propagate
the high-level features from the previous key frame via spatially variant
convolution.


Note that the combined cost of kernel prediction and spatially variant
convolution is dramatically lower than computing the high-level features
from $F_l^t$ ($38$ ms vs $299$ ms).
With our design, such an
adaptive propagation scheme can maintain reasonably high accuracy
for a certain range ($7$ frames) from a key frame.
Moreover, the process of deciding whether $I^t$ is a key frame is
also cheap ($20$ ms).
Hence, by selecting key frames smartly and propagating features
effectively, we can significantly reduce the overall cost.

\vspace{-0.16cm}
\subsection{Adaptive Selection of Key Frames}
\label{sub:keyframe}


An important step in our pipeline is to decide which frames
are the key frames.
A good strategy is to select key frames more frequently
when the video is experiencing rapid changes, while reducing the
computation when the observed scene is stable. Whereas this has been
mentioned in previous literatures, what dominate the practice are still
fixed-rate schedulers~\cite{zhu2017deep} or those based on simple thresholding of
feature variances~\cite{shelhamer2016clockwork}.


According to the rationale above,
a natural criterion for judging whether a frame should be chosen as a new key frame
is the \emph{deviation} of its segmentation map from that of the previous key frame.
This can be formally defined as the fraction of pixels at which
the semantic labels differ.
Intuitively, a large deviation implies significant changes and
therefore it would be the time to set a new key frame.

However, computing the \emph{deviation} as defined above requires
the segmentation map of the current frame, which is expensive to obtain.
Our approach to this problem is to leverage the low-level
features to \emph{predict} its value.
Specifically, we conducted an empirical study on both Cityscapes and Camvid
datasets, and found that there exists strong correlation between the difference
in low-level features and the deviation values.
Greater differences in the low-level features usually indicate larger deviations.

\begin{figure}[t]
    \centering
    \includegraphics[width = 1.0\linewidth]{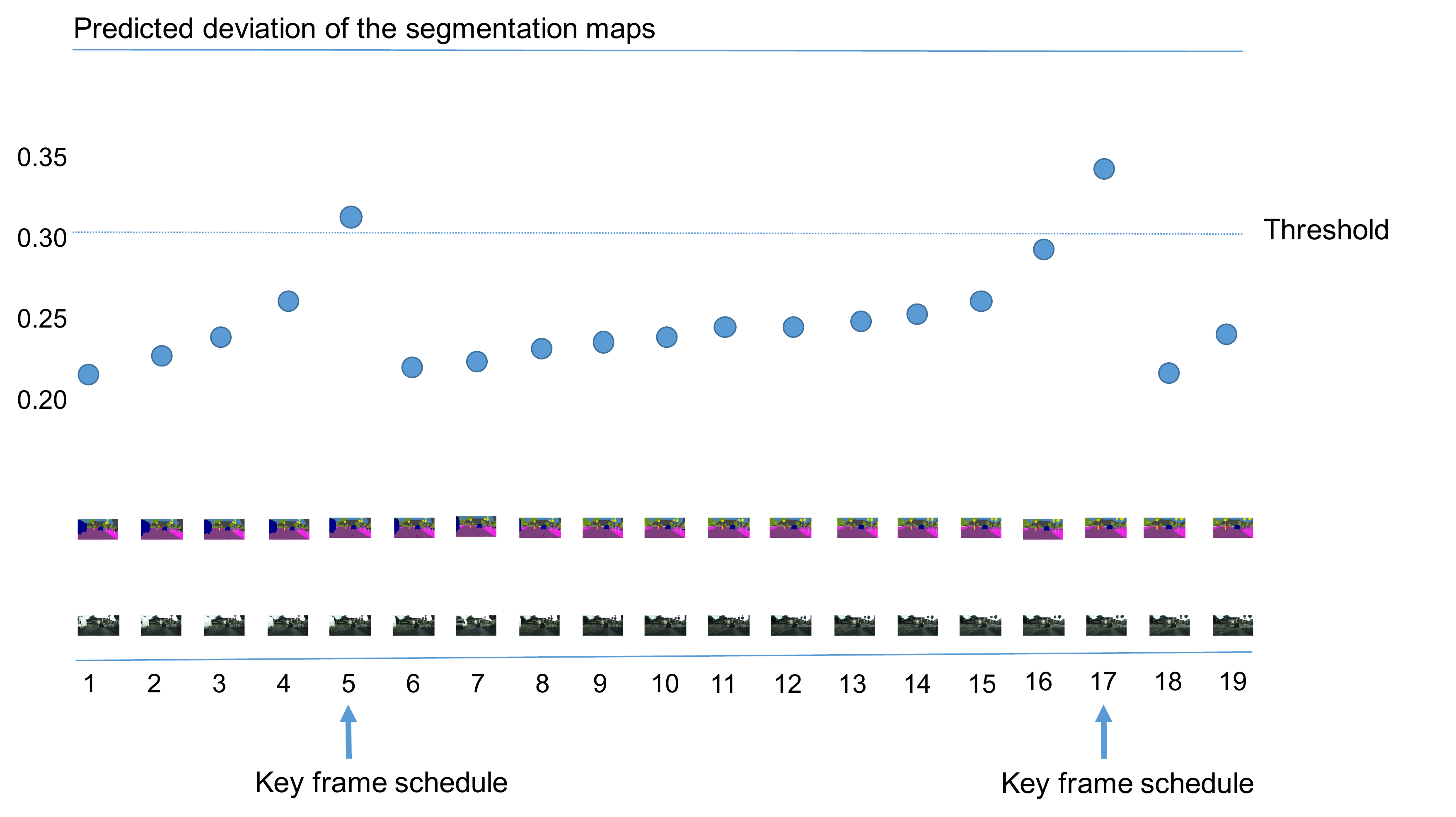}
    \caption{\small
        Adaptive key frame selection.
        As we proceed further away from the key frame, the predicted deviation
        of the segmentation map, depicted with blue dots, gradually increases.
        In our scheme, when the deviation goes beyond
        a pre-defined threshold, the current frame will be selected
        as a new key frame.
    }
    \label{fig:selectkey}
\end{figure}

Motivated by this observation, we devise a small neural network to make
the prediction. Let $k$ and $t$ be the indexes of two frames, this network
takes the differences between their low-level features,
\ie~$(F_l^t - F_l^k)$, as input, and predicts the segmentation deviation,
denoted by $\mathrm{dev}_S(k, t)$.
Specifically, our design of this prediction network comprises
two convolutional kernels with $256$ channels, a global pooling and a fully-connected layer
that follows.
In runtime, at time step $t$, we use this network to predict the
deviation from the previous key frame, after the low-level features are
extracted. As shown in Figure~\ref{fig:selectkey}, we observed that
the predicted deviation would generally increases over time.
If the predicted deviation goes beyond a pre-defined threshold,
we set the current frame as a key frame, and computes its high-level
features with $S_h$, the higher part of the CNN.

\subsection{Adaptive Feature Propagation}
\label{sub:propagate}


As mentioned, for a frame $I^t$ that is not a key frame, its high-level features
will be derived by propagating from the previous key frame,
which we denote by $I^k$.
The question of \emph{how to propagate features effectively and efficiently}
is nontrivial. Existing works often adopt either of the following two approaches.
1) \emph{Follow optical flows}~\cite{zhu2017deep}.
While sounding reasonable, this way has two drawbacks:
(a) The optical flows are expensive to compute.
(b) Point-to-point mapping is often too restrictive.
For high-level feature maps, where each feature actually captures the visual
patterns over a neighborhood instead of a single site, a linear combination
may provide greater latitude to express the propagation more accurately.
2) \emph{Use translation-invariant convolution}~\cite{mahasseni2017budget}.
While convolution is generally less expensive and offers greater flexibility,
using a fixed set of convolution kernels uniformly across the map
is problematic. Different parts of the scene have different motion patterns,
\eg~they may move towards different directions,
therefore they need different weights to propagate.


Motivated by this analysis, we propose to propagate the features
by \emph{spatially variant convolution}, that is, using convolution
to express linear combinations of neighbors, with the kernels varying
across sites. Let the size of the kernels be $H_K \times H_K$,
then the propagation from
the high-level features of the previous key frame ($F_h^k$)
to that of the current frame ($F_h^t$) can be expressed as
{\small
\begin{equation} \label{eq:svconv}
    F_h^t(l, i, j) =
    \sum_{u=-\Delta}^{\Delta} \sum_{v=-\Delta}^{\Delta}
    W^{(k,t)}_{ij}(u, v) \cdot F_h^k(l, i - u, j - v).
\end{equation}
}
Here, $\Delta = \left\lfloor {H_K / 2} \right\rfloor$,
$F_h^t(l, i, j)$ is the feature value at $(i, j)$ of the $l$-th channel
in $F_h^t$,
$W^{(k,t)}_{ij}$ is an $H \times H$ kernel used to compute
the feature at $(i, j)$ when propagating from $F_h^k$ to $F_h^t$.
Note that the kernel values are to assign weights
to different neighbors, which are dependent on the feature location
$(i, j)$ but shared across all channels.

\begin{figure}[t]
    \centering
    \includegraphics[height=250pt]{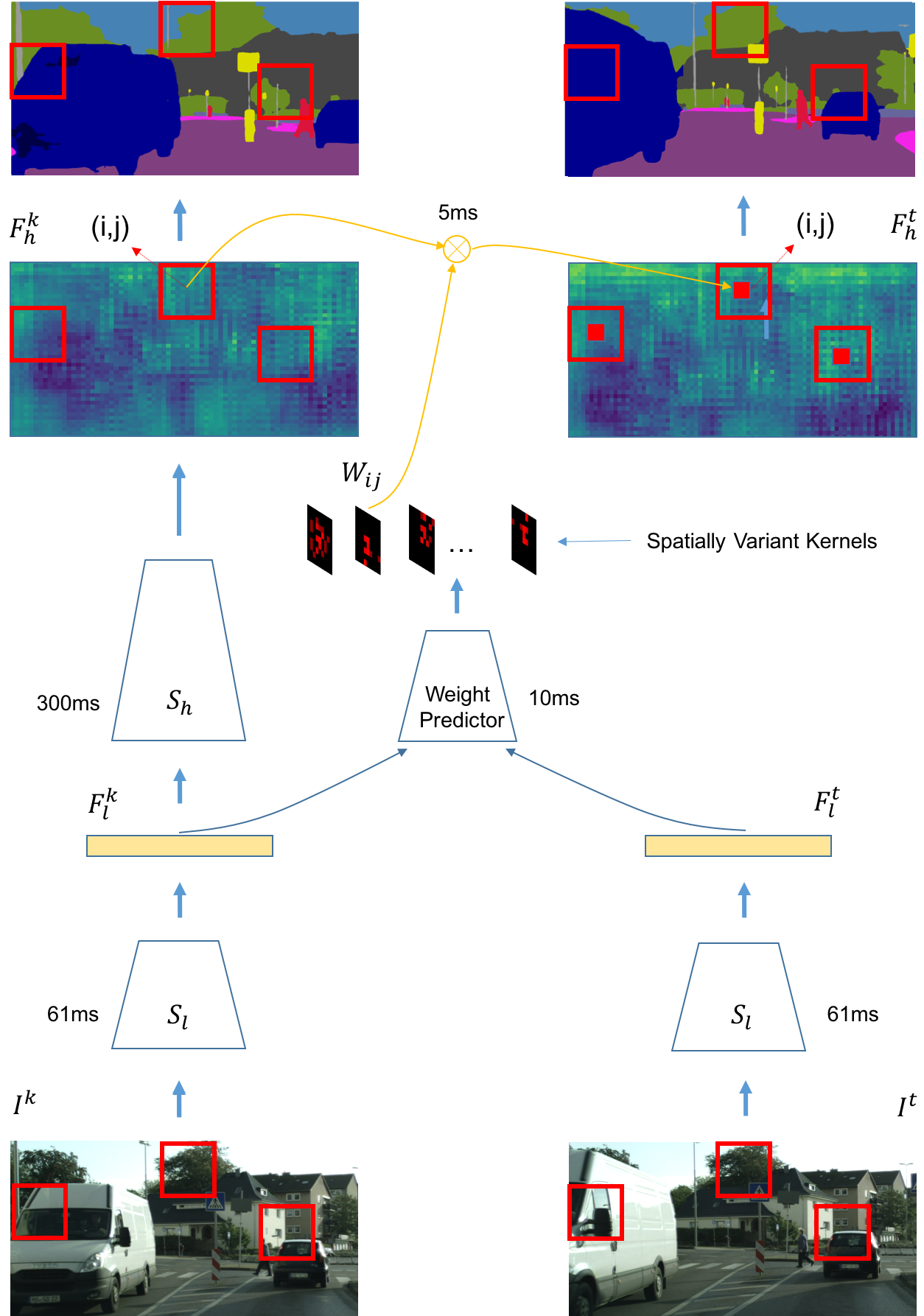}%
    \caption{\small
        Adaptive feature propagation.
        Let $t$ refer to the current frame and $k$ the previous key frame.
        When the low-level features $F_l^t$ are computed,
        the kernel weight predictor will take both $F_l^k$ and $F_l^t$
        as input, and yield a series of convolution kernels, each for
        a different location. Then the high-level features in the previous
        key frame $F_h^k$ will be propagated to the current time step
        via spatially variant convolution using the predicted kernels.
        Finally, the high-level features will be used in pixel-wise label
        prediction.
    }
    \label{propagation}
\end{figure}

There remains a question
-- how to obtain the spatially variant kernels $W_{ij}^{(k,t)}$.
Again, we leverage low-level features to solve the problem.
In particular, we devise a \emph{kernel weight predictor},
which is a small network that takes the low-level features of both
frames, $F_l^k$ and $F_l^t$, as input, and produces the kernels
at all locations altogether.
This network comprises three convolutional layers interlaced with ReLU
layers. The output of the last convolutional layer is of size
$H_K^2 \times H \times W$, where $H \times W$ is the spatial size
of the high-level feature map. This means that it outputs an $H_K^2$-channel
vector at each location, which is then repurposed to be a kernel of size
$H_K \times H_K$ for that location.
This output is then converted to normalized weights via a softmax layer
to ensure that the weights of each kernel sum to one.
With the weights decided on the low-level features, we allow the kernels
to be adapted to not only the locations but also the frame contents,
thus obtaining great expressive power.


To increase the robustness against scene changes, we fuse the low-level
features $F_l^t$ with the propagated high-level feature $F_h^t$
for predicting the labels.
Unlike in original Clockwork Net~\cite{shelhamer2016clockwork},
where low- and high-level features are simply concatenated,
we introduce a low-cost \emph{AdaptNet} to adapt the low-level features
before it is used for prediction.
The \emph{AdaptNet} consists of three convolution layers  and each layer has a small number of channels.
The \emph{AdaptNet} is jointly learned with the weight predictor
in model training. In this way, the framework can learn to exploit the
complementary natures of both components more effectively.

\subsection{Low-Latency Scheduling}
\label{sub:low_latency}


Low latency is very important in many real-world applications,
\eg~surveillance and autonomous driving.
Yet, it has not received much attention in previous works.
While some existing designs~\cite{shelhamer2016clockwork} can reduce the
overall cost in an amortized sense, the maximum latency is not decreased
in this design due to the heavy computation at key frames.

Based on the framework presented above, we devise a new scheduling scheme
that can substantially reduce the maximum latency. The key of our approach
is to introduce a \emph{``fast track''} at key frames.
Specifically, when a frame $I^t$ is decided to be a key frame, this scheme
will compute the segmentation of this frame through the \emph{``fast track''},
\ie~via feature propagation. The high-level features resulted from the
fast track are temporarily treated as the new key frame feature and placed
in the cache.
In the mean time, a \emph{background process} is launched to compute the
more accurate version of $F_h^t$ via the \emph{``slow track''} $S_h$,
\emph{without blocking} the main procedure.
When the computation is done, this version will replace the cached features.

Our experiments show that this design can significantly reduce the maximum
latency (from $360$ ms to $119$ ms), only causing
minor drop in accuracy (from $76.84\%$ to $75.89\%$). Hence, it is a very
effective scheme for real-world applications, especially those
with stringent latency constraints.
Note that the low-latency scheme cannot work in isolation.
This good balance between performance and latency can only be achieved
when the low-latency scheme is working with the adaptive key-frame selection
and the adaptive feature propagation modules presented above.


\section{Implementation Details}
\label{sec:implement}

Our basic network is a ResNet-101~\cite{he2015deep} pretrained on ImageNet.
We choose \texttt{conv4\_3} as the split point between the lower and higher
parts of the network.
The low-level features $F_l^t$ derived from this layer has $1024$ channels.
The lower part consumes about $1/6$ of the total inference time.
In general, the model can achieve higher accuracy
with a heavier low-level part, but at the expense of higher computing cost.
We chose \emph{conv4\_3} as it attains a good tradeoff
between accuracy and speed on the validation sets.

The \emph{adaptive key frame selector} takes the low-level features
of the current frame $F_l^t$ and that of the previous key frame $F_l^k$
as input. This module first reduces the input features
to $256$ channels with a convolution layer with $3 \times 3$ kernels
and then compute their differences, which are subsequently
fed to another convolution layer with
$256$ channels and $3 \times 3$ kernels.
Finally, a global pooling and a fully-connected layer
are used to predict the deviation.

The \emph{kernel weight predictor} in
the \emph{adaptive feature propagation} module (see Fig.~\ref{propagation})
has a similar structure, except that the input features are
concatenated after being reduced to $256$ channels,
the global pooling is removed, and
the fully-connected layer is replaced by a convolution layer with
$1 \times 1$ kernels and $81$ channels.
The \emph{AdaptNet} also reduces the low-level features to $256$ channels by a
convolution layer with $3 \times 3$ kernels, and then pass them to
two two convolution layers with $256$ channels and $3 \times 3$ kernels.
For non-key frames, the adapted low-level features, \ie~the output of the
\emph{AdaptNet} are fused with the propagated high-level features.
The \emph{fusion} process takes the concatenation of both features as input,
and sends it through a convolution layer with $3\times3$ kernels and $256$
channels.

During training, we first trained basic network with respective
ground-truths and fixed it as the feature extractor. Then, we finetuned the adaptive propagation module and the adaptive schedule module, both of which can take a pair of frames
that are $l$ steps apart as input ($l$ is randomly chosen in $[2, 10]$).
Here, we choose the pairs such that
in each pair the first frame is treated as the key frame, and the second
one comes with annotation. For the adaptive propagation module, the kernel predictor and the AdaptNet
are integrated into a network. This integrated network can produce
a segmentation map for the second frame in each pair through
keyframe computation and propagation. This combined network is trained
to minimize the loss between the predicted segmentation (on the second frame
of each pair) and the annotated groundtruth.
For training the adaptive schedule module, we generated the segmentation maps of all unlabelled frames  as auxiliary labels based on the trained basic network and computed the deviation of the segmentation maps between key frame and current frame as the regression target.
The training images were randomly cropped to $713\times713$ pixels. No other data argumentation methods were used.


\section{Experiment}
\label{sec:exper}

We evaluated our framework on two challenging datasets,
Cityscapes~\cite{cordts2016cityscapes} and CamVid~\cite{brostow2009semantic},
and compared it with state-of-the-art approaches.
Our method, with lowest latency, outperforms previous methods significantly.

\subsection{Datasets and Evaluation Metrics}

\emph{Cityscapes}~\cite{cordts2016cityscapes} is set up for urban scene
understanding and autonomous driving. It contains snippets of street scenes
collected from $50$ different cities, at a frame rate of $17$ fps.
The training, validation, and test sets respectively
contain $2975$, $500$, and $1525$ snippets.
Each snippet has $30$ frames, where the $20$-th frame is annotated
with pixel-level ground-truth labels for semantic segmentation with $19$
categories. The segmentation accuracy is measured by the pixel-level
\emph{mean Intersection-over-Union (mIoU)} scores.

\emph{Camvid}~\cite{brostow2009semantic} contains $701$ color images
with annotations of $11$ semantic classes.
These images are extracted from driving videos captured at daytime and dusk.
Each video contains $5000$ frames on average,
with a resolution of $720 \times 960$ pixels.
Totally, there are about $40K$ frames.

\subsection{Evaluation on Cityscapes Dataset}

\begin{table}[tbp]
\centering
\begin{tabular}{c|ccc}
\hline
Method & mIOU & Avg RT & Latency \\
\hline
Clockwork Net~\cite{shelhamer2016clockwork} & 67.7\% & 141ms & 360ms\\
Deep Fea. Flow~\cite{zhu2017deep}           & 70.1\% & 273ms & 654ms \\
GRFP(5)~\cite{nilsson2016semantic}          & 69.4\% & 470ms & 470ms\\
\hline
baseline           & 80.2\%           & 360ms & 360ms\\
AFP + fix schedule & 75.26\%          & 151ms & 360ms\\
AFP + AKS          & \textbf{76.84\%} & 171ms & 380ms\\
AFP + AKS + LLS    & 75.89\%          & 119ms & 119ms\\
\hline
\end{tabular}
\vspace{0.1cm}
\caption{\small
    Comparison with state-of-the-art on Cityscapes dataset.
    ``Avg RT'' means \emph{average runtime per frame}.
    ``AFP'' means \emph{adaptive feature propopagation}.
    ``fix schedule'' means key frame is selected very 5 frame.
    ``AKS'' means \emph{adaptive key frame selection}.
    ``LLS'' means \emph{low-latency scheduling scheme}. Clockwork Net is implmented with the same settings as our method with fix schedule: (a) the same backbone network (ResNet-101),
(b) the same split between the low-level and high-level stages, (c) the same AdaptNet following the low-level stage,
and (d) the same interval between keyframes ($l = 5$).
}\label{tab:cityscapes}
\end{table}

We compared our low-latency video semantic segmentation framework with recent
state-of-the-art methods, following their evaluation protocol.
Table~\ref{tab:cityscapes} shows the quantitative comparison.
The baseline is per-frame segmentation with the full ResNet.
Our adaptive feature propagation with fixed-interval schedule
speeds up the pipeline, reducing
the per-frame runtime from $360$ ms to $151$ ms
while decreasing the performance by $4.9\%$.
Using adaptive key frame selection, the performance is boosted by $1.6\%$.
While these schemes reduce the overall runtime, they are not able
to reduce the maximum latency, due to the heavy computation on key frames.
The low-latency scheduler effectively tackles this issue, which
substantially decreases the latency to $119$ ms (about $1/3$ of the baseline
latency) while maintaining a comparable performance (with just a minor drop).

From the results, we also see that other methods proposed recently
fall short in certain aspects.
In particular, Clockwork Net~\cite{shelhamer2016clockwork} has the same latency, but at the cost of significantly
dropped performance.
DFF~\cite{zhu2017deep} maintains a better performance, but at the expense
of considerably increased computing cost and dramatically larger latency.
Our final result, with low latency schedule for video segmentation, outperforms
previous methods by a large margin. This validates the effectiveness of
our entire design.

\begin{figure*}[ht]
    \centering
    \includegraphics[width=0.96\textwidth]{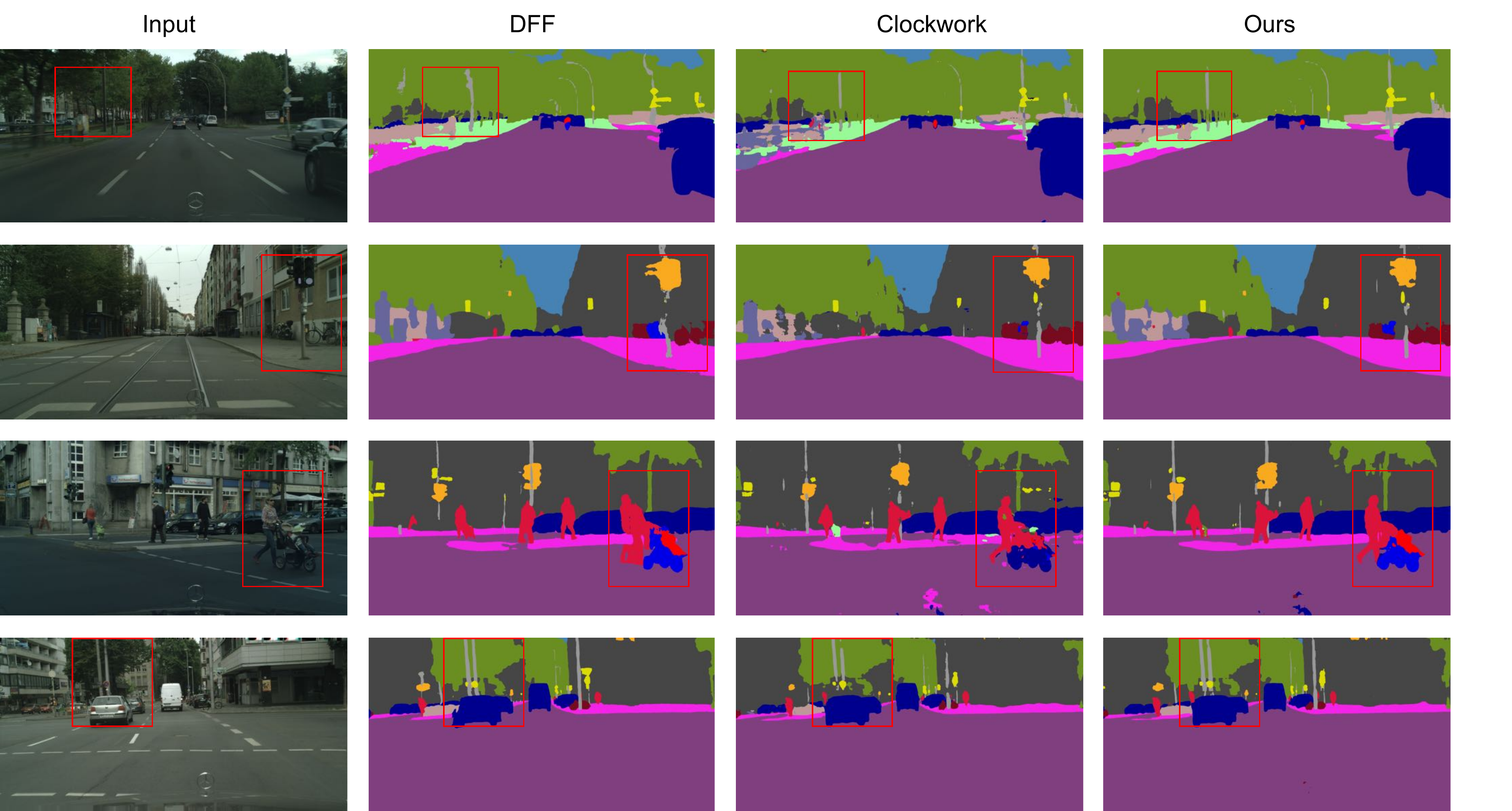}
    \caption{Visual Results for Cityscapes Dataset: our method can achieve significant improvement shown as the red rectangle.}
    \label{fig:cityscapes_show}
\end{figure*}

In addition to the quantitative comparison, we also shows some visual results in
Fig.~\ref{fig:cityscapes_show}. Our method can successfully learn the video
segmentation even when the frames vary significantly. In what follows, we
will investigate the design for each module respectively.

\begin{table}[tbp]
\begin{center}
\begin{tabular}{c|ccc}
\hline
Method & mIOU \\
\hline
Clockwork Propagation~\cite{shelhamer2016clockwork} & 56.53\% \\
Optical Flow Propagation~\cite{zhu2017deep} & 69.2\%\\
\hline
Unified Propagation &	58.73\% \\
Weight By Image Difference &	60.12\% \\
Adaptive Propagation Module (no fuse) &	68.41\% \\
Adaptive Propagation Module (with fuse) &	\textbf{75.26\%} \\
\hline
\end{tabular}
\vspace{0.1cm}
\caption{\small
    Comparison of different feature propagation modules.
}
\label{tab:propagation_fixed}
\end{center}
\end{table}

\begin{figure}[t]
\centering
\includegraphics[width=0.9\linewidth]{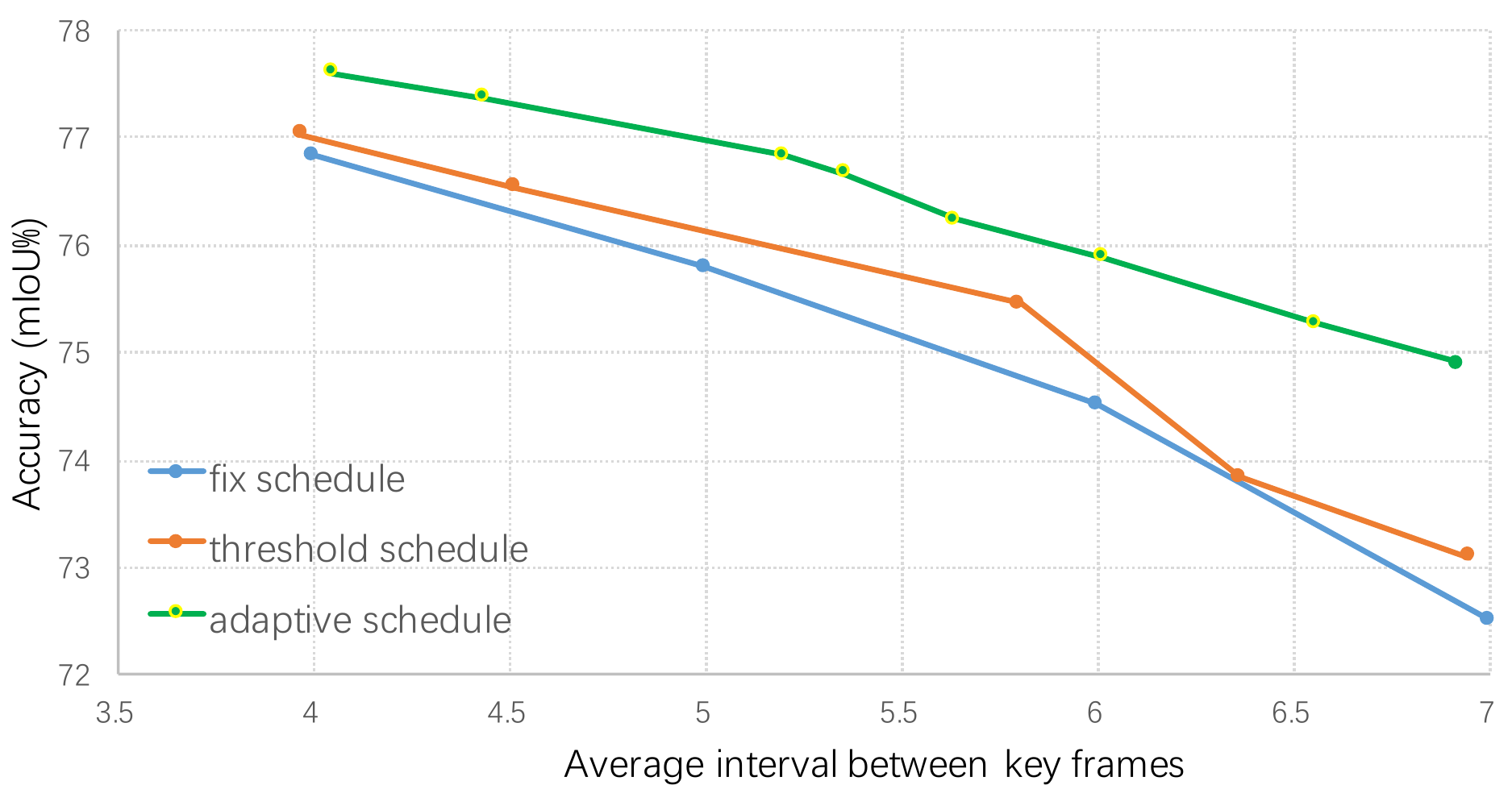}%
 \caption{\small
    Comparison of different scheduling schemes.
 }
 \label{schedule_result}
\end{figure}

\vspace{-7pt}
\paragraph{Comparison on Feature Propagation}

We begin by evaluating the effectiveness of our proposed adaptive propagation
module. To evaluate the specific performance gain on this module, we fix the
scheduling scheme as previous methods, namely selecting a key-frame per $5$
frames.

Table~\ref{tab:propagation_fixed} shows the quantitative comparison.
We compared our method with a globally learned unified propagation, for which the
result is not satisfactory. It reveals that the spatially variant weight is
quite important in our solution.
We also compared to a baseline, where the weights are directly set by
the differences of the input pixel values, our learned propagation weights
also perform better.
Our experiment also shows that the fuse with the low-level features
from the current frame by \emph{AdaptNet} achieves significant
performance improvements (from $68.41\%$ to $75.26\%$), which may be ascribed
to the strong complementary information learned by the fuse model.

Compared to recently proposed methods for feature propagation,
our adaptive propagation module still performs significantly better.
Particularly, Clockwork directly replaces parts of the current features with
previous ones, and therefore the derived representation is not adapted to
current changes, thus resulting in poor performance.
The optical flow based method DFF~\cite{zhu2017deep} relies on the optical
flows to propagate features and shows substantially better accuracies
as compared to Clockwork, due to its better adaptivity.
Yet, its performance is sensitive to the quality of the Flownet, and
it ignores the spatial relationship on the feature space.
These factors limit its performance gain, and hence it is still inferior
to the proposed method.

\vspace{-7pt}
\paragraph{Comparison on Scheduling}
\label{exp_adaptive_schedule}

We also studied the performance of the adaptive scheduling module for
key frame selection, given our feature propagation module.
Note that a good schedule not only reduces the frequency of key frames,
but also increases the performance given a fixed number of key frames.
Hence, we conducted several experiments to compare different schedule methods
across different key-frame intervals.

Specifically, we compared
fixed rate schedule, threshold schedule, and also our adaptive schedule module.
Fig.~\ref{schedule_result} shows the results.
Under each key-frame interval, our adaptive schedule always outperforms
the other two.
We believe the reason why the proposed method outperforms the
threshold schedule is that the intermediate
feature maps is not specially optimized for key-frame selection.
They may contain noises and perhaps many other irrelevant factors.

\begin{table}[tbp]
\centering
{
\begin{tabular}{c|ccc}
\hline
Modules & Time (ms) & R \\
\hline
Lower part of network $S_l$  &61 & 16.9\%\\
Higher part of network $S_h$  &299  & 83.1\%\\
Basic Network &360 & 100\% \\
\hline
Adaptive Schedule Module  &20 & 5.5\%\\
Adaptive Propagation Module &38 & 10.5\%\\
\hline
Non-Key Frame &119 & 33\%\\
\hline
\end{tabular}
}
\vspace{0.1cm}
\caption{Latency analysis for each module of our network.
The second column (R) shows the ratio of the latency
to the time spent by the basic network.
}
\label{tab:latency}
\end{table}

\vspace{-7pt}
\paragraph{Cost and Latency Analysis}
We analyzed the computation cost of each component in our pipeline, and
then the overall latency of our method. The results are in Table~\ref{tab:latency}.
With the low-latency scheduling scheme,
the total latency of our pipeline is equal to the sum
of the lower part of the network $S_l$,
adaptive key-frame selection, adaptive feature propagation,
which is $0.119s$ ($33\%$ of the basic network).
On the contrary, all previous methods fail to reduce latency in
their system design, despite that they may reduce the overall cost
in the amortized sense. Hence, they are not able to meet the latency requirements
in real-time applications.




\begin{table}[hbp]
\centering
\begin{tabular}{c|ccc}
\hline
Methods & Pixel Accuracy & CA\\
\hline
SuperParsing~\cite{tighe2010superparsing}  & 83.9\% & 62.5\%\\
DAG-RNN~\cite{shuai2016dag}  &91.6\%  &78.1\%\\
MPF-RNN~\cite{jin2016multi}  &92.8\%  &82.3\%\\
RTDF~\cite{lei2016recurrent}  &89.9\%  &80.5\%\\
PEARL~\cite{jin2016video}  &94.2\%  &82.5\%\\
Ours & \textbf{94.6\%} &\textbf{82.9}\%\\
\hline
\end{tabular}
\vspace{0.1cm}
\caption{\small
    Result comparison for CamVid dataset.
}
\label{tab:camvid}
\end{table}

\begin{figure*}[ht]
    \centering
    \includegraphics[width=0.96\textwidth]{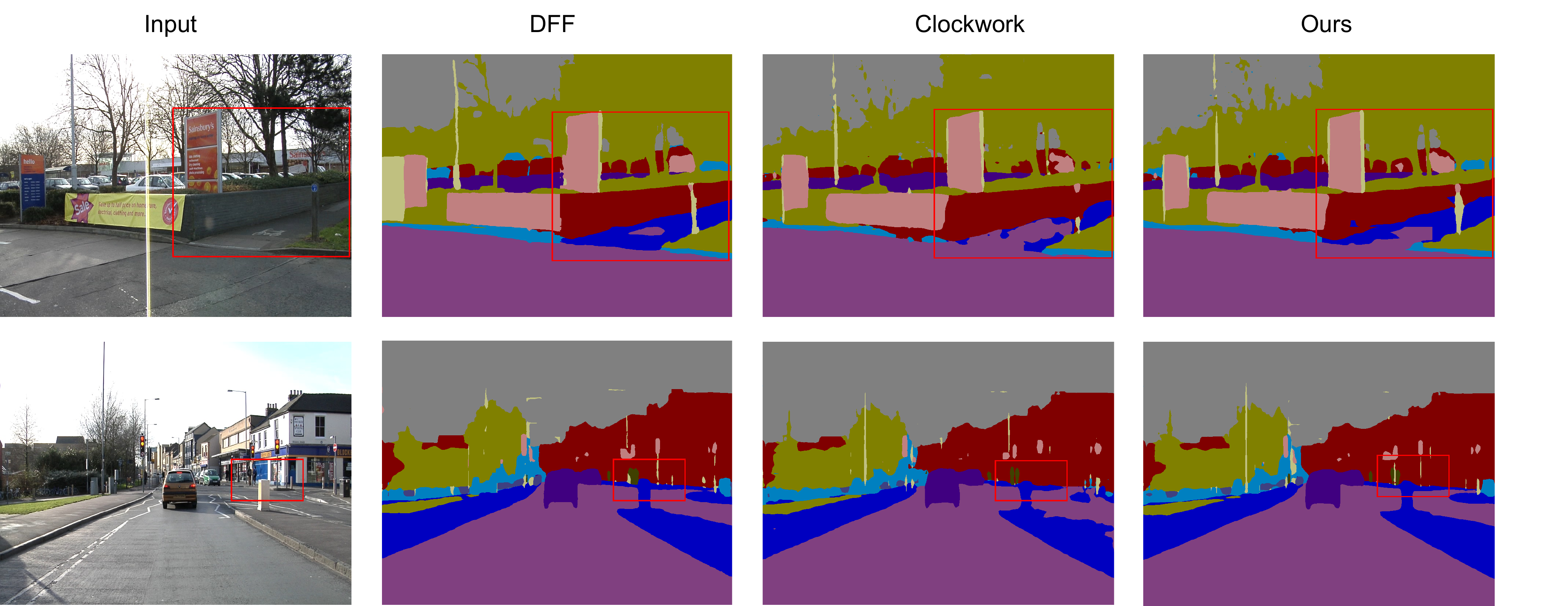}
    \caption{Visual Results for Camvid Dataset: our method can achieve significant improvement shown as the red rectangle.}
    \label{fig:camvid_show} \vspace{-0.3cm}
\end{figure*}

\subsection{CamVid Dataset}

We also evaluated our method on CamVid, another video segmentation dataset,
and compard it with multiple previous methods that have reported performances
on CamVid, which range from traditional methods to various CNN or RNN based methods.
In Table~\ref{tab:camvid},
we report the pixel accuracy and average per-Class Accuracy(CA),
which can reduce the dominated effect on majority classes.
We can see our framework still performs the best.
The consistently good performances on both datasets show that
our adaptive framework for video segmentation method is in general beneficial
to video perception.
Qualitative results on this dataset are shown in Fig.~\ref{fig:camvid_show}.



\section{Conclusion}
We presented an efficient video semantic segmentation framework
with two key components:
adaptive feature propagation and adaptive key-frame schedule.
Particularly, our specially designed schedule scheme can achieve low latency
in an online setting.
The results on both Cityscapes and CamVid showed that
our method can yield a substantially better tradeoff between accuracy
in latency, compared to previous methods.
In future, we will
explore more model compression approaches which can further reduce the overall
computation cost and latency for a practical system.

\section*{Acknowledgements}


This work was partially supported by the Big Data Collaboration Research grant from SenseTime Group (CUHK Agreement No. TS1610626), the Early Career Scheme (ECS) of Hong Kong (No. 24204215), and the 973 Program under contract No. 2015CB351802.

{\small
\bibliographystyle{ieee}
\bibliography{egbib}

\begin{thebibliography}{10}\itemsep=-1pt

\bibitem{badrinarayanan2015segnet}
V.~Badrinarayanan, A.~Kendall, and R.~Cipolla.
\newblock Segnet: A deep convolutional encoder-decoder architecture for image
  segmentation.
\newblock {\em CoRR}, abs/1511.00561, 2015.

\bibitem{brostow2009semantic}
G.~J. Brostow, J.~Fauqueur, and R.~Cipolla.
\newblock Semantic object classes in video: A high-definition ground truth
  database.
\newblock {\em Pattern Recognition Letters}, 30(2):88--97, 2009.

\bibitem{caelles2016one}
S.~Caelles, K.-K. Maninis, J.~Pont-Tuset, L.~Leal-Taix{\'e}, D.~Cremers, and
  L.~Van~Gool.
\newblock One-shot video object segmentation.
\newblock {\em arXiv preprint arXiv:1611.05198}, 2016.

\bibitem{chen2014semantic}
L.~Chen, G.~Papandreou, I.~Kokkinos, K.~Murphy, and A.~L. Yuille.
\newblock Semantic image segmentation with deep convolutional nets and fully
  connected crfs.
\newblock {\em CoRR}, abs/1412.7062, 2014.

\bibitem{chen2015attention}
L.~Chen, Y.~Yang, J.~Wang, W.~Xu, and A.~L. Yuille.
\newblock Attention to scale: Scale-aware semantic image segmentation.
\newblock {\em CoRR}, abs/1511.03339, 2015.

\bibitem{chen2016deeplab}
L.-C. Chen, G.~Papandreou, I.~Kokkinos, K.~Murphy, and A.~L. Yuille.
\newblock Deeplab: Semantic image segmentation with deep convolutional nets,
  atrous convolution, and fully connected crfs.
\newblock {\em arXiv preprint arXiv:1606.00915}, 2016.

\bibitem{cordts2016cityscapes}
M.~Cordts, M.~Omran, S.~Ramos, T.~Rehfeld, M.~Enzweiler, R.~Benenson,
  U.~Franke, S.~Roth, and B.~Schiele.
\newblock The cityscapes dataset for semantic urban scene understanding.
\newblock In {\em Proceedings of the IEEE Conference on Computer Vision and
  Pattern Recognition}, pages 3213--3223, 2016.

\bibitem{fayyaz2016stfcn}
M.~Fayyaz, M.~H. Saffar, M.~Sabokrou, M.~Fathy, R.~Klette, and F.~Huang.
\newblock Stfcn: Spatio-temporal fcn for semantic video segmentation.
\newblock {\em arXiv preprint arXiv:1608.05971}, 2016.

\bibitem{fischer2015flownet}
P.~Fischer, A.~Dosovitskiy, E.~Ilg, P.~H{\"a}usser, C.~Haz{\i}rba{\c{s}},
  V.~Golkov, P.~van~der Smagt, D.~Cremers, and T.~Brox.
\newblock Flownet: Learning optical flow with convolutional networks.
\newblock {\em arXiv preprint arXiv:1504.06852}, 2015.

\bibitem{gadde2017semantic}
R.~Gadde, V.~Jampani, and P.~V. Gehler.
\newblock Semantic video cnns through representation warping.
\newblock {\em arXiv preprint arXiv:1708.03088}, 2017.

\bibitem{hariharan2015hypercolumns}
B.~Hariharan, P.~A. Arbel{\'{a}}ez, R.~B. Girshick, and J.~Malik.
\newblock Hypercolumns for object segmentation and fine-grained localization.
\newblock In {\em CVPR}, pages 447--456, 2015.

\bibitem{he2015deep}
K.~He, X.~Zhang, S.~Ren, and J.~Sun.
\newblock Deep residual learning for image recognition.
\newblock {\em CoRR}, abs/1512.03385, 2015.

\bibitem{jin2016multi}
X.~Jin, Y.~Chen, J.~Feng, Z.~Jie, and S.~Yan.
\newblock Multi-path feedback recurrent neural network for scene parsing.
\newblock {\em arXiv preprint arXiv:1608.07706}, 2016.

\bibitem{jin2016video}
X.~Jin, X.~Li, H.~Xiao, X.~Shen, Z.~Lin, J.~Yang, Y.~Chen, J.~Dong, L.~Liu,
  Z.~Jie, et~al.
\newblock Video scene parsing with predictive feature learning.
\newblock {\em arXiv preprint arXiv:1612.00119}, 2016.

\bibitem{khoreva2016learning}
A.~Khoreva, F.~Perazzi, R.~Benenson, B.~Schiele, and A.~Sorkine-Hornung.
\newblock Learning video object segmentation from static images.
\newblock {\em arXiv preprint arXiv:1612.02646}, 2016.

\bibitem{krizhevsky2012imagenet}
A.~Krizhevsky, I.~Sutskever, and G.~E. Hinton.
\newblock Imagenet classification with deep convolutional neural networks.
\newblock In {\em Advances in neural information processing systems}, pages
  1097--1105, 2012.

\bibitem{lei2016recurrent}
P.~Lei and S.~Todorovic.
\newblock Recurrent temporal deep field for semantic video labeling.
\newblock In {\em European Conference on Computer Vision}, pages 302--317.
  Springer, 2016.

\bibitem{liu2015semantic}
Z.~Liu, X.~Li, P.~Luo, C.~C. Loy, and X.~Tang.
\newblock Semantic image segmentation via deep parsing network.
\newblock In {\em ICCV}, pages 1377--1385, 2015.

\bibitem{long2015fully}
J.~Long, E.~Shelhamer, and T.~Darrell.
\newblock Fully convolutional networks for semantic segmentation.
\newblock In {\em CVPR}, pages 3431--3440, 2015.

\bibitem{mahasseni2017budget}
B.~Mahasseni, S.~Todorovic, and A.~Fern.
\newblock Budget-aware deep semantic video segmentation.
\newblock In {\em Proceedings of the IEEE Conference on Computer Vision and
  Pattern Recognition}, pages 1029--1038, 2017.

\bibitem{nilsson2016semantic}
D.~Nilsson and C.~Sminchisescu.
\newblock Semantic video segmentation by gated recurrent flow propagation.
\newblock {\em arXiv preprint arXiv:1612.08871}, 2016.

\bibitem{noh2015learning}
H.~Noh, S.~Hong, and B.~Han.
\newblock Learning deconvolution network for semantic segmentation.
\newblock In {\em ICCV}, pages 1520--1528, 2015.

\bibitem{paszke2016enet}
A.~Paszke, A.~Chaurasia, S.~Kim, and E.~Culurciello.
\newblock Enet: {A} deep neural network architecture for real-time semantic
  segmentation.
\newblock {\em CoRR}, abs/1606.02147, 2016.

\bibitem{perazzi2016benchmark}
F.~Perazzi, J.~Pont-Tuset, B.~McWilliams, L.~Van~Gool, M.~Gross, and
  A.~Sorkine-Hornung.
\newblock A benchmark dataset and evaluation methodology for video object
  segmentation.
\newblock In {\em Proceedings of the IEEE Conference on Computer Vision and
  Pattern Recognition}, pages 724--732, 2016.

\bibitem{shelhamer2016clockwork}
E.~Shelhamer, K.~Rakelly, J.~Hoffman, and T.~Darrell.
\newblock Clockwork convnets for video semantic segmentation.
\newblock In {\em Computer Vision--ECCV 2016 Workshops}, pages 852--868.
  Springer, 2016.

\bibitem{shuai2016dag}
B.~Shuai, Z.~Zuo, B.~Wang, and G.~Wang.
\newblock Dag-recurrent neural networks for scene labeling.
\newblock In {\em Proceedings of the IEEE Conference on Computer Vision and
  Pattern Recognition}, pages 3620--3629, 2016.

\bibitem{simonyan2014very}
K.~Simonyan and A.~Zisserman.
\newblock Very deep convolutional networks for large-scale image recognition.
\newblock {\em CoRR}, abs/1409.1556, 2014.

\bibitem{tighe2010superparsing}
J.~Tighe and S.~Lazebnik.
\newblock Superparsing: scalable nonparametric image parsing with superpixels.
\newblock In {\em European conference on computer vision}, pages 352--365.
  Springer, 2010.

\bibitem{zhao2016pyramid}
H.~Zhao, J.~Shi, X.~Qi, X.~Wang, and J.~Jia.
\newblock Pyramid scene parsing network.
\newblock {\em arXiv preprint arXiv:1612.01105}, 2016.

\bibitem{zheng2015conditional}
S.~Zheng, S.~Jayasumana, B.~Romera{-}Paredes, V.~Vineet, Z.~Su, D.~Du,
  C.~Huang, and P.~H.~S. Torr.
\newblock Conditional random fields as recurrent neural networks.
\newblock In {\em ICCV}, pages 1529--1537, 2015.

\bibitem{zhu2017deep}
X.~Zhu, Y.~Xiong, J.~Dai, L.~Yuan, and Y.~Wei.
\newblock Deep feature flow for video recognition.
\newblock In {\em Proceedings of the IEEE Conference on Computer Vision and
  Pattern Recognition}, pages 2349--2358, 2017.

\end{thebibliography}
}

\end{document}